\documentclass{article} 
\usepackage{iclr2027_conference,times}

\usepackage{amsmath,amsfonts,bm}

\def\eqref#1{equation~\ref{#1}}

\def\1{\bm{1}}

\DeclareMathAlphabet{\mathsfit}{\encodingdefault}{\sfdefault}{m}{sl}
\SetMathAlphabet{\mathsfit}{bold}{\encodingdefault}{\sfdefault}{bx}{n}

\usepackage{hyperref}
\usepackage{url}
\usepackage{amsmath,amsfonts}
\usepackage[ruled,vlined]{algorithm2e}
\usepackage{array}
\usepackage[caption=false,font=normalsize,labelfont=sf,textfont=sf]{subfig}
\usepackage{fontawesome5}
\usepackage{textcomp}
\usepackage{stfloats}
\usepackage{verbatim}
\usepackage{graphicx}
\usepackage{booktabs}
\usepackage{tcolorbox}
\usepackage{listings}
\usepackage{xcolor}
\usepackage{wrapfig}
\usepackage{graphicx}
\usepackage{tabularx}
\usepackage{makecell}
\usepackage{enumitem}
\usepackage{pifont}
\usepackage{multirow}
\usepackage[table]{xcolor}
\hypersetup{
    colorlinks=true,
    citecolor=darkgray,
    linkcolor=blue,
    urlcolor=blue!55!black
}

\title{Mental-R1: Aligning LLM Reasoning for Mental Health Assessment}

\title{Mental-R1: Aligning LLM Reasoning for Mental Health Assessment}

\author{
Xin Wang$^{1}$ \quad
Boyan Gao$^{1}$ \quad
Yibo Yang$^{2}$ \quad
David A. Clifton$^{1,3}$ \\
$^{1}$Department of Engineering Science, University of Oxford, Oxford, U.K. \\
$^{2}$School of Artificial Intelligence, Shanghai Jiao Tong University, Shanghai, China \\
$^{3}$Oxford Suzhou Centre for Advanced Research, Suzhou, China \\
\texttt{xin.wang@eng.ox.ac.uk} \quad
\texttt{david.clifton@eng.ox.ac.uk}
}

\iclrfinalcopy 
\begin{document}

\maketitle

\lhead{}
\begin{abstract}
Mental health problems such as anxiety, depression, and suicide remain urgent global challenges, where timely and accurate assessment is critical for effective intervention. Recently, large language models have been explored for mental health assessment. However, existing general-purpose post-training methods do not align with the cognitive processes of human assessment, which may lead to unreliable reasoning outcomes.
To bridge this gap, we propose Cognitive Relative Policy Optimization (CRPO), a reinforcement learning framework tailored for the mental health domain. CRPO extends group relative policy optimization by integrating stage-dependent uncertainty modeling into the policy optimization process. Specifically, we introduce a stage-wise entropy regularization mechanism that encourages broad exploration in early reasoning phases and progressively enforces confident decision-making in later stages, mimicking the human cognitive shift from uncertainty to certainty. 
In addition, inspired by cognitive appraisal theory, we formalize cognitive reasoning stages, thereby guiding theory-grounded interpretable inference.
Experiments on 8 mental health datasets show that CRPO achieves an average improvement of 10.4 percentage points in weighted F1-score over the best reinforcement learning baseline. Furthermore, the CRPO-trained model Mental-R1 demonstrates clear advantages compared with existing large language models on reasoning-intensive cases, suggesting that CRPO enhances reasoning capabilities for mental health assessment. The project repository is available at
\url{https://github.com/xin-wang18/Mental-R1}.
\end{abstract}

\section{Introduction}

Mental health problems such as depression and suicidal behavior have become global burdens for human well-being. According to the World Health Organization, more than 720,000 people die by suicide each year, which is equivalent to one person every 43 seconds~\citep{WHO2023Suicide}. Early assessment of mental health conditions is therefore crucial for timely intervention and prevention. Mental Health Assessment (MHA) focuses on identifying individuals' mental conditions, including loneliness~\citep{jiang2022many,wang2024decoding}, depression~\citep{sampath2022data,naseem2022early}, stress~\citep{wang2020leverage,wang2022meta}, anxiety~\citep{owen2020towards,yu2023automatic}, and suicide risk~\citep{cao2019latent,cao2021learning} from their textual statements.

Recently, large language models (LLMs) have emerged as a promising paradigm for mental health assessment~\citep{shi2025mentalqlm,xu2024mental,yang2024mentallama,hu2025pattern,ravenda2025llms,rohei2026review,zhai2025mentalglm} due to their strong generalization ability across diverse natural language understanding tasks~\citep{naveed2025comprehensive,zhao2025cyberconfucius,hu2025beyond,hu2025emobench,chen2026llm}. Prior studies primarily use standard supervised fine-tuning (SFT)~\citep{gao2025optimization} or reinforcement learning (RL)~\citep{kumar2025llm} to adapt LLMs for mental health tasks. However, these ``generalist'' post-training methods often fail to reflect the real-world assessment process of mental health professionals, limiting their reliability in healthcare applications.

In real-world assessment, mental health professionals often seek to understand an individual's condition by reconstructing their underlying cognitive process~\citep{beck2020cognitive,persons2012case,kuyken2011collaborative,eells2022handbook}.
Two characteristics of this reasoning process are particularly important. First, the assessment process follows the natural dynamics of human cognition~\citep{elstein1978medical,higgs2024clinical,garb1998studying}.
At early stages, mental health professionals tend to collect observations and explore possible signals with relatively high uncertainty.
As more contextual information is considered, they gradually refine their interpretation and move toward a more confident assessment.
This uncertainty-to-certainty transition reflects a fundamental property of human cognitive reasoning~\citep{gold2007neural,clark2013whatever}. Second, such assessments typically follow theory-grounded cognitive stages~\citep{wright2017learning}.
Consistent with the cognitive-behavioral framework and the ABC model~\citep{ellis1962reason,beck1979cognitive}, the process often begins by identifying potential stimuli or life events that may contribute to psychological distress.
It then analyzes how the individual cognitively appraises these events~\citep{lazarus1984stress} and how such appraisals lead to affective or behavioral reactions.
Finally, the individual's mental state is inferred.

Inspired by these characteristics of cognitive process reconstruction, we propose Cognitive Relative Policy Optimization (CRPO), a reinforcement learning framework that aligns LLM reasoning with human cognitive dynamics and theory-grounded cognitive stages.

To model the cognitive dynamics, we introduce stage-wise entropy regularization, a stage-dependent uncertainty control mechanism integrated into the policy optimization objective. Unlike many reinforcement learning approaches~\citep{Remax,reinforce++,guo2025deepseek} that apply a uniform exploration strategy throughout the reasoning process, our method explicitly modulates entropy across reasoning stages. Early reasoning stages are encouraged to maintain higher entropy to promote diverse exploration, while later stages gradually reduce entropy to guide the model toward more confident conclusions. This design enables a principled stage-aware exploration-conclusion trade-off that mirrors the uncertainty-to-certainty transition in human cognition.

To capture the theory-grounded reasoning stages, we draw inspiration from cognitive appraisal theory~\citep{lazarus1984stress,ellsworth1991some,watson2007causes}, a classic psychological framework that explains the internal cognitive processes underlying human mental responses. Based on this theory, we formalize a set of reasoning stages, including stimulus, primary appraisal, secondary appraisal, reaction, and mental state.
We operationalize this framework during training by designing a format reward that encourages outputs consistent with these reasoning stages.

In summary, CRPO bridges the reasoning gap between generalist LLM post-training methods and real-world mental health assessment practice,
thereby improving performance. The main contributions of this paper are threefold:
\begin{itemize}

\item \textbf{Cognitive-Inspired Reinforcement Learning.}
We propose Cognitive Relative Policy Optimization (CRPO), a reinforcement learning framework that aligns LLM reasoning with human cognitive dynamics. Our primary algorithmic contribution is stage-wise entropy regularization, which translates the ``uncertainty-to-certainty'' cognitive shift into a stage-dependent uncertainty modulation mechanism within the policy optimization objective. Furthermore, we formalize theory-grounded reasoning stages inspired by cognitive appraisal theory to support interpretable inference.

\item \textbf{Extensive Empirical Validation.}
Experiments on 8 mental health datasets show that CRPO consistently outperforms existing post-training baselines, yielding an average improvement of 10.4 percentage points in weighted F1-score. 
Furthermore, the CRPO-trained model Mental-R1 demonstrates an advantage of approximately 15.6 percentage points over the best-performing LLM on complex samples, suggesting that CRPO effectively enhances model's reasoning capabilities for mental health.

\item \textbf{Transparent Benchmark.}
The evaluation of mental health assessment can be hindered by the mixed accessibility of existing benchmarks, where datasets are not uniformly open. To facilitate this area, we construct a transparent benchmark based entirely on publicly accessible datasets. By systematically comparing modern RL-based and LLM baselines, we provide a solid foundation for advancing 
interdisciplinary research in AI and healthcare.

\end{itemize}
\section{Related Work}
\paragraph{Mental Health Assessment.}
Mental health assessment in computational research has primarily focused on identifying conditions such as depression~\citep{sampath2022data,naseem2022early,fisher2026language}, stress~\citep{wang2023contrastive,wang2025mise,ikae2026scoping}, anxiety~\citep{owen2020towards,yu2023automatic,hidayat2025language}, and suicide risk~\citep{cao2019latent,kina2026suicide} from textual data. These tasks are typically formulated as classification problems.
Recent studies have applied large language models (LLMs) to mental health assessment
~\citep{yulianti2025feelings,nanda2024detecting,jin2025applications}.
\citet{xu2024mental} evaluated multiple LLMs on textual mental health prediction tasks, showing that their instruction-tuned models such as Alpaca and FLAN-T5 substantially outperform prompt-based baselines. \citet{yang2024mentallama} enhanced interpretability by constructing an explanation dataset through ChatGPT generation, and fine-tuned LLaMA-2 to jointly improve prediction and explanation quality.
\citet{shi2025mentalqlm} proposed a lightweight 0.5B-parameter model with dual LoRA modules and data pruning, achieving competitive results on benchmark datasets with much lower resource requirements.
However, these works do not align with human cognitive dynamics and lack theory-grounded staged reasoning.
To address these limitations, our work explicitly introduces a stage-wise entropy regularization strategy that guides LLM reasoning from early-stage exploration to final certainty, and integrates cognitive appraisal theory to define structured reasoning stages.

\paragraph{Reinforcement Learning for Reasoning.}
Reinforcement learning (RL) has recently become a central paradigm for enhancing the reasoning ability of large language models~\citep{lightman2023let,shao2024deepseekmath,guo2025deepseek,covo2025,yue2025does,zhang2025survey}. A representative approach is reinforcement learning from human feedback, i.e., RLHF~\citep{ouyang2022training}, which fine-tunes models with preference data to align their outputs with human judgments.
In parallel, Direct Preference Optimization, i.e., DPO~\citep{rafailov2023direct} reformulates preference learning into a supervised objective, enabling more stable and efficient optimization.
Beyond preference learning, recent work has explored how RL can directly improve reasoning quality. Group Relative Policy Optimization, i.e., GRPO~\citep{guo2025deepseek} has been proposed as an efficient alternative for training LLMs on reasoning tasks, using group-based relative rewards to enhance sample efficiency and stability. Other studies further extend this line of research:
RL Tango~\citep{tango2025} employs joint generator-verifier training to bolster robustness.
DAPO~\citep{yu2025dapo} introduces decoupled clipping strategies alongside dynamic sampling to optimize training stability and signal quality.
However, these approaches do not reflect how human cognition works, where early reasoning tends to be exploratory and later reasoning becomes more decisive. Our CRPO framework builds on this line by introducing a stage-wise entropy regularization strategy, which explicitly modulates exploration and certainty across reasoning stages, thereby aligning LLM reasoning more closely with human cognition.

\begin{table*}[t]
\tiny
\centering
\setlength{\tabcolsep}{3.7pt}
\caption{Comparison of our CRPO-trained Mental-R1 with existing mental health–focused LLMs. Mental-R1 algorithmically introduces cognition-aligned uncertainty dynamics and theory-grounded cognitive reasoning stages.}
\label{tab:model_comparison}

\begin{tabular}{lccccccccc}
\toprule
\textbf{Model} 
& \textbf{Training} 
& \textbf{Uncertainty Dynamics} 
& \textbf{Response Trajectory} 
& \textbf{Benchmark Datasets} 
& \textbf{Str} & \textbf{Anx} & \textbf{Dep} & \textbf{Sui} & \textbf{Lon} \\
\midrule

Mental-QLM~\citep{shi2025mentalqlm} 
& SFT & Not modeled & Answer + explanation & 5 public
& \ding{51} & \ding{55} & \ding{51} & \ding{55} & \ding{55} \\

Mental-LLM~\citep{xu2024mental} 
& SFT & Not modeled & Answer + explanation & 5 public
& \ding{51} & \ding{55} & \ding{51} & \ding{51} & \ding{55} \\

Mental-Llama~\citep{yang2024mentallama} 
& SFT & Not modeled & Answer + explanation & 5 public
& \ding{51} & \ding{55} & \ding{51} & \ding{51} & \ding{51} \\

Mental-GLM~\citep{zhai2025mentalglm} 
& SFT & Not modeled & Answer + explanation & 2 public
& \ding{55} & \ding{55} & \ding{55} & \ding{51} & \ding{55} \\

\midrule
Mental-R1 (Ours) 
& CRPO (RL) 
& \makecell[c]{Cognition-aligned\\exploration $\rightarrow$ certainty} 
& \makecell[c]{Cognitive stages\\+ answer} 
& \makecell[c]{8 public} 
& \ding{51} & \ding{51} & \ding{51} & \ding{51} & \ding{51} \\

\bottomrule
\end{tabular}

\vspace{2pt}
\footnotesize{
Str: Stress; Anx: Anxiety; Dep: Depression; 
Sui: Suicide; Lon: Loneliness.
}
\end{table*}

\section{Cognitive Relative Policy Optimization (CRPO)}
\begin{figure*}
  \centering
  \includegraphics[width=1\columnwidth]{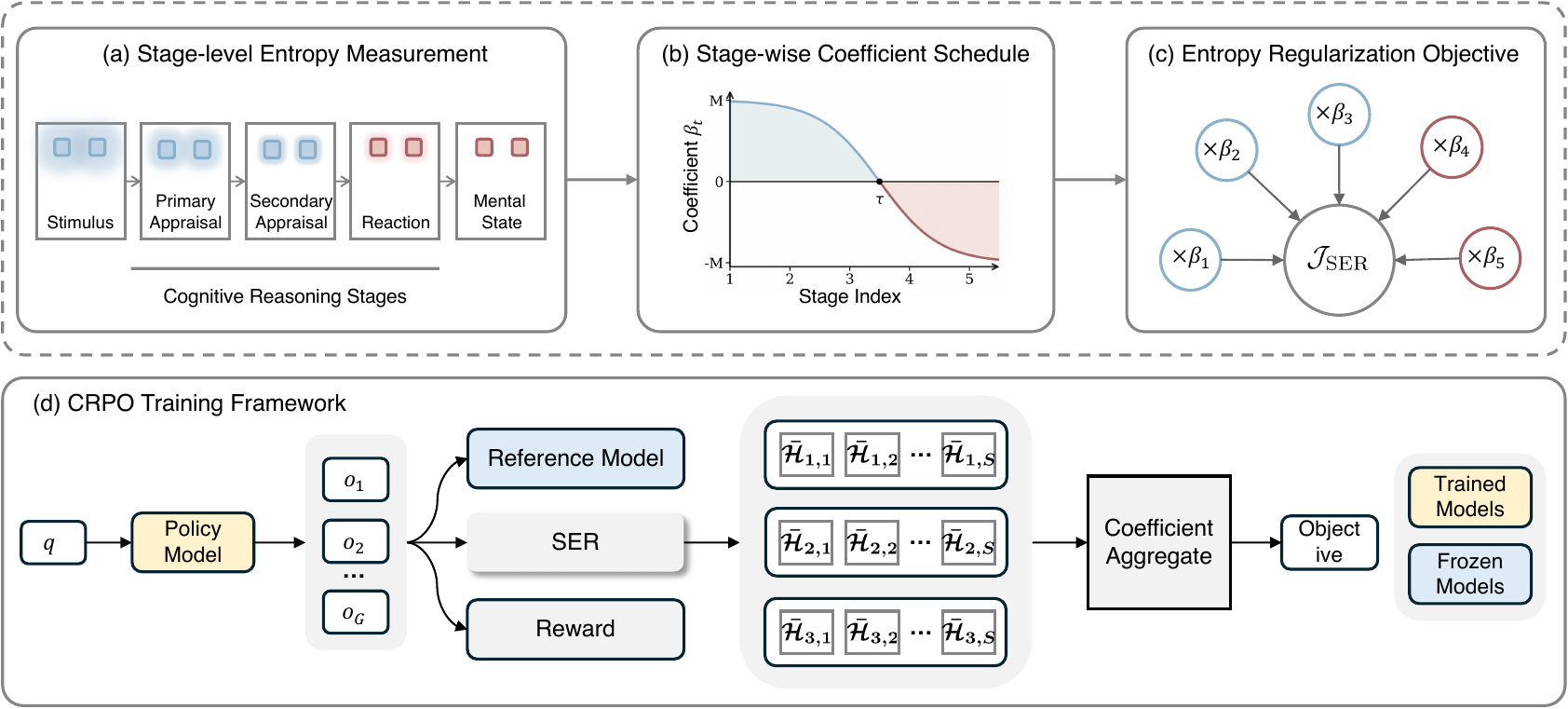}
  \caption{Bottom: Overall illustration of the proposed CRPO reinforcement learning framework. The objective is derived from reward-based advantages and the Stage-wise Entropy Regularization (SER) module, with a reference model used for stability. Top: Detailed illustration of SER. The process begins by measuring stage-level entropy for each reasoning stage. This entropy is then modulated by stage-wise coefficients derived from a hyperbolic tangent schedule. Finally, the stage-level entropy and stage-wise coefficients are combined to form the SER objective. 
  }
  \label{fig:framework}
\end{figure*}

In this section, we present Cognitive Relative Policy Optimization (CRPO), a reinforcement learning framework specifically designed for mental health assessment. An overview of the proposed CRPO framework is illustrated in Figure~\ref{fig:framework}. 

\subsection{Preliminary}
CRPO is built on Group Relative Policy Optimization (GRPO), a critic-free reinforcement learning framework designed for reasoning-oriented generation that estimates advantages via relative comparisons among a group of sampled outputs~\citep{guo2025deepseek}.
This relative optimization paradigm provides a robust foundation for our CRPO, upon which stage-aware uncertainty regularization can be incorporated.
Formally, for each prompt $q$, a group of $G$ outputs $\{o_1, o_2, \dots, o_G\}$ is sampled from the old policy $\pi_{\theta_{old}}$. Each $o_i$ corresponds to one independently generated completion sampled from the same prompt. The training objective to be maximized is defined as:
\begin{equation}
\begin{aligned}
\mathcal{J}_{\mathrm{GRPO}}(\theta) = \frac{1}{G} \sum_{i=1}^G \frac{1}{|o_i|} \sum_{k=1}^{|o_i|} \Bigl[ \min\bigl( \rho_{i,k}(\theta) A_i,
\text{clip}(\rho_{i,k}(\theta), 1-\epsilon, 1+\epsilon) A_i \bigr) - \lambda D_{KL}(\pi_{\theta} \| \pi_{\text{ref}}) \Bigr].
\end{aligned}
\end{equation}
Here, $\rho_{i,k}(\theta)$ is the current-to-old policy probability ratio for token $o_{i,k}$ conditioned on $(q,o_{i,<k})$, and $A_i$ is the reward of completion $o_i$ standardized by the group reward mean and standard deviation. The KL divergence is evaluated at the same token context against a fixed reference policy $\pi_{\mathrm{ref}}$. The coefficients $\epsilon$ and $\lambda$ control clipping and KL regularization, respectively.

\subsection{Stage-wise Entropy Regularization}
While GRPO provides a stable optimization foundation, it treats all reasoning stages uniformly and therefore does not align with the uncertainty evolution of human multi-stage reasoning. Cognitive theory suggests that certainty rarely remains constant across stages; instead, reasoning typically begins with exploratory uncertainty and gradually develops into confident conclusions~\citep{gold2007neural,clark2013whatever}. To address this limitation, we propose Stage-wise Entropy Regularization (SER), which modulates the model's uncertainty across the reasoning stages.
The guiding idea is intuitive: allow high uncertainty in early reasoning stages, and gradually encourage lower uncertainty and greater confidence in later stages, where clarity and decisiveness are required.

\textbf{Stage-level entropy.}
SER modulates uncertainty across reasoning stages to mimic the human uncertainty evolution process. To this end, we quantify the policy's uncertainty within each reasoning stage using a stage-level entropy. For a generated output $o_i$ and reasoning stage $t$, the stage-level entropy is defined as the average entropy over all tokens belonging to that stage:
\begin{equation}
    \bar{\mathcal{H}}_{i,t}(\theta) = \frac{1}{|\mathcal{N}_{i,t}|} \sum_{k \in \mathcal{N}_{i,t}} \mathcal{H}_{i,k}(\theta),
    \label{eq:stage_entropy}
\end{equation}
where $i$ indexes a generated output in the sampled group, $\mathcal{N}_{i,t}$ is the set of token positions in $o_i$ that belong to stage $t$, and $|\mathcal{N}_{i,t}|$ is the number of tokens assigned to that stage. Here, $\mathcal{H}_{i,k}(\theta)$ denotes the token entropy at position $k$:
\begin{equation}
    \mathcal{H}_{i,k}(\theta) = -\sum_{w \in V} \pi_\theta(w \mid q, o_{i,<k}) \log \pi_\theta(w \mid q, o_{i,<k}),
    \label{eq:token_entropy}
\end{equation}
where $w$ is a token in the vocabulary $V$, $q$ denotes the input prompt, and $o_{i,<k}$ is the partial sequence of $o_i$ generated before position $k$.

The stage-level entropy characterizes the policy's uncertainty within a reasoning stage, with higher values indicating more exploratory reasoning and lower values indicating more confident reasoning.
Normalizing by $|\mathcal{N}_{i,t}|$ removes the influence of variable stage lengths, so that the contribution of each stage to the regularization term is governed by its stage-wise coefficient rather than by how many tokens it contains.

Based on the stage-level entropy, the overall entropy regularization term is defined as a weighted sum over all reasoning stages and all generated outputs:
\begin{equation}
\mathcal{J}_{\text{SER}}(\theta)
= \frac{1}{G} \sum_{i=1}^{G}
\sum_{t=1}^{S} \beta_t \, \bar{\mathcal{H}}_{i,t}(\theta),
\end{equation}
where $G$ denotes the number of generated outputs sampled for each prompt, $S$ is the total number of reasoning stages, and $\beta_t$ is a stage-specific coefficient that controls both the magnitude and direction of entropy regularization at stage $t$. The outer average over $G$ ensures that the regularization term is applied consistently across all sampled completions, while the inner summation aggregates the contributions from different reasoning stages. This formulation aligns the uncertainty regularization process with the group-based optimization paradigm of GRPO, treating each generated completion as an independent carrier of stage-wise uncertainty information.

\textbf{Stage-wise coefficient schedule.}
To achieve the intended uncertainty-to-certainty progression, we set $\beta_t$ according to a hyperbolic tangent schedule:
\begin{equation}
    \beta_t=-M \cdot \frac{e^{t-\tau}-e^{-(t-\tau)}}{e^{t-\tau}+e^{-(t-\tau)}},
\end{equation}
where $M$ controls the maximum magnitude of the entropy regularization term, $t$ denotes the reasoning stages index in the cognitive reasoning stages, and $\tau$ specifies the transition point where $\beta_t$ changes sign, i.e., the stage where the regularization term shifts from positive values to negative values. This design provides a smooth progression: 
\begin{itemize}
    \item Early stages ($t < \tau$): receive positive $\beta_t$ to increase entropy, thereby encouraging diverse exploration.
    \item Later stages ($t > \tau$): receive negative $\beta_t$ to suppress entropy, thereby enforcing more deterministic reasoning.
\end{itemize}

\textbf{Overall objective.}
The overall training objective to be maximized is a combination of group relative policy optimization objective and the stage-wise entropy regularization objective:
\begin{equation}
\begin{aligned}
\mathcal{J}_{\text{CRPO}}(\theta) = &\frac{1}{G}\sum_{i=1}^{G} \frac{1}{|o_i|} \sum_{k=1}^{|o_i|} \Big[ \min\big(\rho_{i,k}(\theta) A_i, \text{clip}(\rho_{i,k}(\theta), 1-\epsilon, 1+\epsilon)A_i\big) - \lambda D_{KL}(\pi_\theta \| \pi_{\text{ref}}) \Big] \\
& + \frac{1}{G}\sum_{i=1}^{G} \sum_{t=1}^{S} \beta_t \Bigg( \frac{1}{|N_{i,t}|} \sum_{k \in N_{i,t}} \bigg[ -\sum_{w \in V}\pi_\theta(w \mid q, o_{i,<k}) \log \pi_\theta(w \mid q, o_{i,<k}) \bigg] \Bigg).
\end{aligned}
\end{equation}

We summarize the overall training procedure in Algorithm~\ref{alg:crpo}.

\subsection{Cognitive Reasoning Stages}
SER operates on a multi-stage reasoning trajectory and therefore requires an explicit decomposition of reasoning into different stages. 
To this end, CRPO imposes a structured set of Cognitive Reasoning Stages inspired by cognitive appraisal theory~\citep{lazarus1984stress,ellsworth1991some, watson2007causes}, a foundational framework in psychology that explains how individuals construct mental states through subjective evaluations of potential factors.
The central principle of this theory is that mental states emerge through staged cognitive appraisals, rather than directly from stimuli.
When confronted with a stimulus, a person engages in a sequence of appraisal processes: first, assessing the nature and significance of the event, and second, evaluating whether sufficient resources are available to cope with it. These appraisals in turn shape the individual’s affective and behavioral responses. An illustrative example is provided in Appendix~\ref{app:cognitive_appraisal_example}.
Building on this theory, we design a structured set of cognitive reasoning stages:
\begin{itemize}[label=--, leftmargin=*]
    \item Stimulus stage: Identify the root stimulus, i.e., the event, situation, or object explicitly or implicitly described in the input text.  
    \item Primary appraisal stage: Infer the individual’s initial evaluation of the stimulus, such as whether it is threatening, positive, or irrelevant.  
    \item Secondary appraisal stage: Determine whether the individual perceives sufficient resources or coping ability to handle the situation.  
    \item Reaction stage: Summarize the likely affective and behavioral responses elicited by the appraisals.  
    \item Mental state stage: Conclude the probable mental health condition or state (e.g., stress, depression, or anxiety) resulting from the preceding reasoning process. 
\end{itemize}
We denote the stages as $c=\langle c_t\rangle_{t=1}^{S}$ with $S=5$. To ensure this reasoning structure in LLM outputs, we design a system prompt and a format reward integrated into reinforcement learning.

\subsection{Reward Design}

The format reward encourages the output to follow the cognitive reasoning stages.
We assign $r_f(o)=+\alpha$ if the following conditions are met: 1) the output contains the outer tags `$<$think$>$' and `$<$answer$>$' exactly once each, and in the correct order; 2) within `$<$think$>$', the stage tags of $c$ appear exactly once each, in the predefined order, and with non-empty content; and 3) the content within the `$<$answer$>$' section is non-empty. Otherwise, $r_f(o)=-\alpha$.

The answer reward encourages the final prediction to match the ground truth.
We assign $r_a(o,y)=+1$ if the predicted answer matches the ground truth $y$, and $-1$ otherwise.

\emph{Combined reward.}
For a completion $o$ associated with dataset $d$ and ground-truth label $y$, the total reward is defined as a weighted sum of the two rewards:
\begin{equation}
r(o,d,y)
=
r_f(o)
+
\eta_{d,y}\,r_a(o,y),
\label{eq:reward}
\end{equation}
where $\eta_{d,y}>0$ controls the contribution of answer correctness. We set
$\eta_{d,y}=\sqrt{w_{d,y}w_d}$, where
\begin{equation}
w_{d,y}
=
\frac{1/n_{d,y}}
{\frac{1}{C_d}\sum_{j=1}^{C_d}(1/n_{d,j})},
\qquad
w_d
=
\frac{1/n_d}
{\frac{1}{D}\sum_{m=1}^{D}(1/n_m)}.
\label{eq:reward_coefficients}
\end{equation}
Here, $n_{d,y}$ is the number of training samples with label $y$ in dataset $d$, $n_d$ is the training size of dataset $d$, $C_d$ is its number of classes, and $D$ is the number of datasets. These coefficients are computed from training-set statistics and held fixed, with the square root smoothing extreme values. 
Note that $\eta_{d,y}$ matters only when both format and answer rewards vary within a group, where correct answers outrank format compliance iff $\eta_{d,y} > \alpha$; this prioritizes answer correctness for rarer classes and smaller datasets in early training. Otherwise, being constant within a group, it cancels under group-wise advantage normalization and becomes inactive once format compliance stabilizes.

\section{Experiment}
We evaluate our method through the following questions. Details are included in Appendix~\ref{app:details}.

Q1: How does CRPO compare with other RL methods, and how effective are its key components?

Q2: How does CRPO-trained Mental-R1 compare with other LLMs?

Q3: How does CRPO improve the model’s reasoning ability?

Q4: How does the SER affect entropy and its schedule affect performance?

\subsection{Datasets}
We adopt 8 different human-annotated datasets to comprehensively evaluate our method, each corresponding to a unique mental health assessment task, as summarized in Table~\ref{tab:datasets}.
Stress state prediction: Dreaddit~\citep{turcan2019dreaddit}. Anxiety state prediction: DATD~\citep{owen2020towards}. Depression level classification: LT-EDI~\citep{sampath2022data}. Depression severity prediction:  DepSeverity~\citep{naseem2022early}. Suicidal ideation detection: SDCNL~\citep{haque2021deep}. Suicide risk severity classification: RSD~\citep{zheng2025rsd}. 
Loneliness state prediction: FIG~\citep{jiang2022many}.
Loneliness intensity prediction: LID~\citep{LID}.
To the best of our knowledge, this evaluation includes the most extensive collection of publicly accessible human-annotated mental health assessment datasets compared with prior work, offering a more comprehensive and transparent benchmark for future research.

\begin{table*}[]
\centering
\tiny
\caption{Comparison with different critic-free RL methods and ablation results (Weighted F1-score, mean $\pm$ std, in \%). Our CRPO outperforms SFT and RL training baselines across all eight datasets.}
\label{tab:rl_results}
\setlength{\tabcolsep}{2.6pt}
\begin{tabular}{l|cccccccc}
\toprule
& DATD & RSD & DepSeverity & LT-EDI & SDCNL & Dreaddit & FIG & LID \\
Method   & {\scriptsize Anxiety D1} & {\scriptsize Suicide D2} & {\scriptsize Depression D3} & {\scriptsize Depression D4} & {\scriptsize Suicide D5} & {\scriptsize Stress D6} & {\scriptsize Loneliness D7} & {\scriptsize Loneliness D8} \\
\midrule
\rowcolor[HTML]{EFEFEF}
\multicolumn{9}{l}{\textit{Optimize methods}}\\
SFT      & $57.13_{\pm0.76}$ & $30.50_{\pm1.14}$ & $59.41_{\pm0.68}$ & $44.89_{\pm1.46}$ & $48.67_{\pm3.92}$ & $73.46_{\pm1.44}$ & $80.61_{\pm1.64}$ & $18.06_{\pm1.77}$ \\
RLOO    & $57.06_{\pm2.03}$ & $22.34_{\pm2.04}$ & $63.67_{\pm2.08}$ & $38.77_{\pm1.68}$ & $50.38_{\pm1.22}$ & $76.34_{\pm1.05}$ & $86.26_{\pm0.58}$ & $17.24_{\pm1.41}$ \\
ReMax    & $57.42_{\pm1.16}$ & $32.19_{\pm2.83}$ & $74.73_{\pm0.21}$ & $49.24_{\pm0.98}$ & $46.07_{\pm0.79}$ & $62.36_{\pm0.57}$ & $85.77_{\pm1.06}$ & $17.85_{\pm3.65}$ \\
Reinforce++ & $49.42_{\pm2.13}$ & $21.81_{\pm1.68}$ & $75.95_{\pm0.18}$ & $47.61_{\pm0.85}$ & $49.94_{\pm0.71}$ & $66.06_{\pm0.75}$ & $86.01_{\pm1.20}$ & $18.68_{\pm4.36}$ \\
DAPO     & $58.26_{\pm1.29}$ & $26.19_{\pm3.34}$ & $72.28_{\pm2.92}$ & $49.69_{\pm0.91}$ & $52.26_{\pm0.86}$ & $73.78_{\pm1.13}$ & $84.62_{\pm1.03}$ & $21.66_{\pm1.17}$ \\ 
GRPO     & $54.26_{\pm1.68}$ & $28.34_{\pm3.69}$ & $75.88_{\pm1.16}$ & $47.40_{\pm1.00}$ & $53.05_{\pm1.42}$ & $72.60_{\pm0.88}$ & $84.91_{\pm1.01}$ & $19.37_{\pm1.84}$ \\ 
\midrule
\rowcolor[HTML]{EFEFEF}
\multicolumn{9}{l}{\textit{Ablation}}\\
w/o SER    & $56.31_{\pm2.60}$ & $42.05_{\pm2.54}$ & $76.78_{\pm0.46}$ & $53.97_{\pm1.76}$ & $56.93_{\pm1.20}$ & $75.39_{\pm1.07}$ & $87.46_{\pm0.52}$ & $29.42_{\pm1.36}$ \\
w/o CRS   & $59.67_{\pm1.57}$ & $46.78_{\pm3.30}$ & $77.65_{\pm1.91}$ & $54.10_{\pm1.67}$ & $60.27_{\pm1.18}$ & $75.98_{\pm1.28}$ & $88.32_{\pm1.53}$ & $34.38_{\pm4.27}$ \\
CRPO  & $\mathbf{63.62_{\pm1.03}}$ & $\mathbf{50.14_{\pm4.72}}$ & $\mathbf{78.86_{\pm1.16}}$ & $\mathbf{57.14_{\pm0.97}}$ & $\mathbf{61.75_{\pm1.40}}$ & $\mathbf{80.56_{\pm0.57}}$ & $\mathbf{91.57_{\pm0.61}}$ & $\mathbf{38.39_{\pm1.66}}$ \\ 
\bottomrule
\end{tabular}
\end{table*}

\begin{table*}[]
\centering
\tiny
\caption{Comparison with LLMs (Weighted F1-score, mean $\pm$ std, in \%). CRPO-trained Mental-R1 outperforms all the baseline LLMs across eight mental health assessment datasets.}
\label{tab:main_results}
\setlength{\tabcolsep}{2.3pt}
\begin{tabular}{l|cccccccc}
\toprule
& DATD & RSD & DepSeverity & LT-EDI & SDCNL & Dreaddit & FIG & LID \\
Model   & {\scriptsize Anxiety D1} & {\scriptsize Suicide D2} & {\scriptsize Depression D3} & {\scriptsize Depression D4} & {\scriptsize Suicide D5} & {\scriptsize Stress D6} & {\scriptsize Loneliness D7} & {\scriptsize Loneliness D8} \\
\midrule
\rowcolor[HTML]{EFEFEF}
\multicolumn{9}{l}{\textit{Domain and public models}}\\
Mentallama    & $53.78_{\pm1.79}$ & $22.88_{\pm5.33}$ & $41.95_{\pm1.54}$ & $21.49_{\pm1.05}$ & $48.50_{\pm2.12}$ & $75.33_{\pm1.03}$ & $81.30_{\pm1.01}$ & $14.28_{\pm1.49}$ \\ 
Mental-GLM    & $59.89_{\pm1.28}$ & $37.51_{\pm2.45}$ & $53.63_{\pm0.33}$ & $30.36_{\pm1.30}$ & $55.79_{\pm1.26}$ & $76.22_{\pm0.36}$ & $80.40_{\pm0.50}$ & $18.88_{\pm1.30}$ \\ 
Gemma-2-SFT   & $54.33_{\pm3.33}$ & $29.48_{\pm3.08}$ & $56.01_{\pm0.68}$ & $40.23_{\pm1.69}$ & $49.82_{\pm0.77}$ & $73.08_{\pm0.43}$ & $83.58_{\pm0.53}$ & $20.33_{\pm0.83}$ \\ 
Llama-3.1-SFT   & $50.78_{\pm3.24}$ & $26.52_{\pm2.89}$ & $63.67_{\pm1.43}$ & $39.92_{\pm1.66}$ & $45.17_{\pm0.82}$ & $70.90_{\pm0.68}$ & $77.46_{\pm0.77}$ & $15.82_{\pm1.95}$ \\ 
\rowcolor[HTML]{EFEFEF}
\multicolumn{9}{l}{\textit{Industry-level models}}\\
DeepSeek-V3   & $52.21_{\pm1.82}$ & $41.99_{\pm3.57}$ & $49.14_{\pm0.69}$ & $35.60_{\pm1.48}$ & $48.53_{\pm1.04}$ & $71.61_{\pm1.14}$ & $78.23_{\pm1.34}$ & $11.84_{\pm1.51}$ \\
DeepSeek-R1   & $56.91_{\pm2.43}$ & $43.16_{\pm3.98}$ & $62.57_{\pm0.55}$ & $26.40_{\pm1.22}$ & $52.75_{\pm0.98}$ & $72.97_{\pm1.22}$ & $84.56_{\pm1.48}$ & $11.63_{\pm0.93}$ \\
GPT-3.5       & $52.78_{\pm2.91}$ & $29.87_{\pm3.83}$ & $50.47_{\pm1.52}$ & $22.36_{\pm2.45}$ & $42.33_{\pm1.32}$ & $68.46_{\pm1.06}$ & $75.41_{\pm1.69}$ & $20.99_{\pm0.56}$ \\
GPT-4o        & $46.88_{\pm2.31}$ & $35.69_{\pm3.70}$ & $52.74_{\pm0.97}$ & $33.23_{\pm2.16}$ & $51.52_{\pm1.18}$ & $71.38_{\pm0.76}$ & $74.51_{\pm0.99}$ & $11.26_{\pm1.38}$ \\
GPT-5         & $56.98_{\pm1.28}$ & $43.08_{\pm1.65}$ & $62.19_{\pm0.79}$ & $36.18_{\pm0.84}$ & $51.18_{\pm1.73}$ & $72.86_{\pm1.84}$ & $81.38_{\pm0.96}$ & $22.68_{\pm0.85}$ \\ 
\midrule
Mental-R1     & $\mathbf{63.62_{\pm1.03}}$ & $\mathbf{50.14_{\pm4.72}}$ & $\mathbf{78.86_{\pm1.16}}$ & $\mathbf{57.14_{\pm0.97}}$ & $\mathbf{61.75_{\pm1.40}}$ & $\mathbf{80.56_{\pm0.57}}$ & $\mathbf{91.57_{\pm0.61}}$ & $\mathbf{38.39_{\pm1.66}}$ \\ 
\bottomrule
\end{tabular}
\end{table*}

\subsection{Q1: Comparison with other RL methods} 
We compare CRPO with supervised fine-tuning and a diverse set of reinforcement learning baselines, which include critic-free methods RLOO~\citep{RLOO}, ReMax~\citep{Remax}, Reinforce++~\citep{reinforce++}, GRPO~\citep{guo2025deepseek}, and DAPO~\citep{yu2025dapo}. 
As reported in the upper portion of Table~\ref{tab:rl_results}, CRPO attains the highest performance across all eight datasets. Averaged across datasets, CRPO outperforms the strongest reinforcement learning baseline DAPO by 10.4 points in Weighted F1, and improves over SFT by 13.6 points, respectively. The gains are especially pronounced on the RSD and LID datasets. These results substantiate the effectiveness and generality of our CRPO framework for mental-health–oriented reasoning tasks. 

Ablation results are presented in the lower portion of Table~\ref{tab:rl_results}. Removing SER leads to the largest average performance drop, confirming the critical role of SER in aligning the reasoning process with human cognitive dynamics, thereby improving overall model performance. 
For the w/o Cognitive Reasoning Stages (CRS) variant, we remove the five cognitive stages to allow free-form reasoning. The reasoning content is divided into $S=5$ contiguous segments of approximately equal length, and SER is applied to these segments with the same schedule. The resulting performance decline supports the benefit of explicit cognitive stage structure over free-form reasoning.

\begin{figure*}[t]
  \centering
  \includegraphics[width=\textwidth]{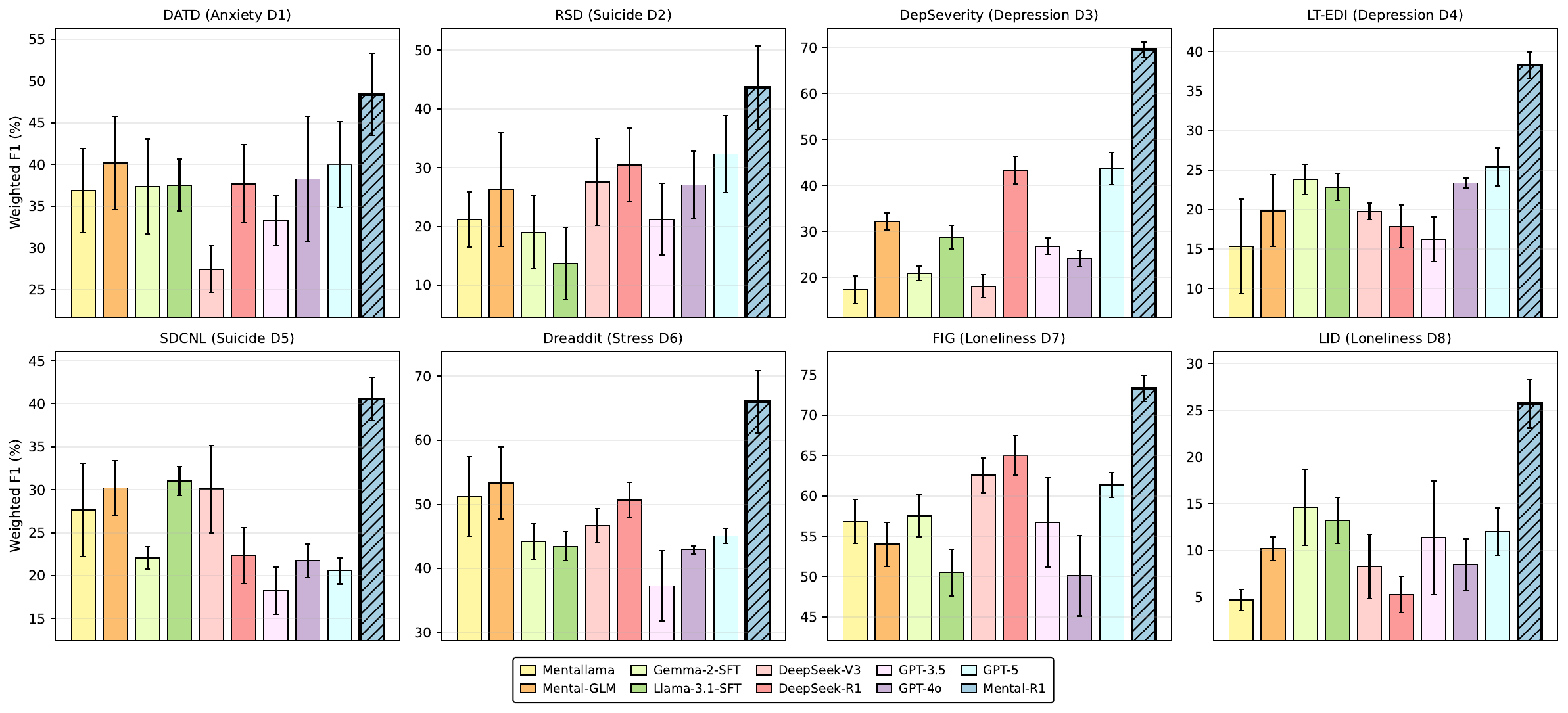}
  \caption{Reasoning-focused evaluation results (Weighted F1-score, mean $\pm$ std, in \%). CRPO-trained Mental-R1 consistently outperforms all LLM baselines on reasoning-intensive samples.}
  \label{fig:reasoning_bars}
\end{figure*}
\subsection{Q2: Comparison with LLMs}
We conduct a comprehensive comparison between CRPO-trained Mental-R1 and three groups of LLM baselines: mental health–specific large language models such as Mentallama-13B~\citep{yang2024mentallama} and Mental-GLM~\citep{zhai2025mentalglm}, open-source LLMs including Gemma-2-9B~\citep{team2024gemma} and Llama-3.1-8B~\citep{dubey2024llama} that are supervised fine-tuned on the same training data for fairness, and large proprietary models such as DeepSeek-V3~\citep{liu2024deepseek}, DeepSeek-R1~\citep{guo2025deepseek}, GPT-3.5~\citep{GPT35}, GPT-4o~\citep{GPT4o}, and GPT-5~\citep{GPT5}. 
The results in Table~\ref{tab:main_results} show that Mental-R1 consistently achieves best performance across all eight datasets. 
On average, Mental-R1 outperforms the best mental health–specific baseline by 13.6 points, open-source baseline by 14.4 points, and proprietary baseline by 11.9 points in weighted-F1.
Mental-R1 delivers particularly strong gains on the LT-EDI depression detection dataset. These results demonstrate that our method enhances the model’s ability for mental health assessment.

\begin{figure*}[]
  \centering
  \includegraphics[width=\textwidth]{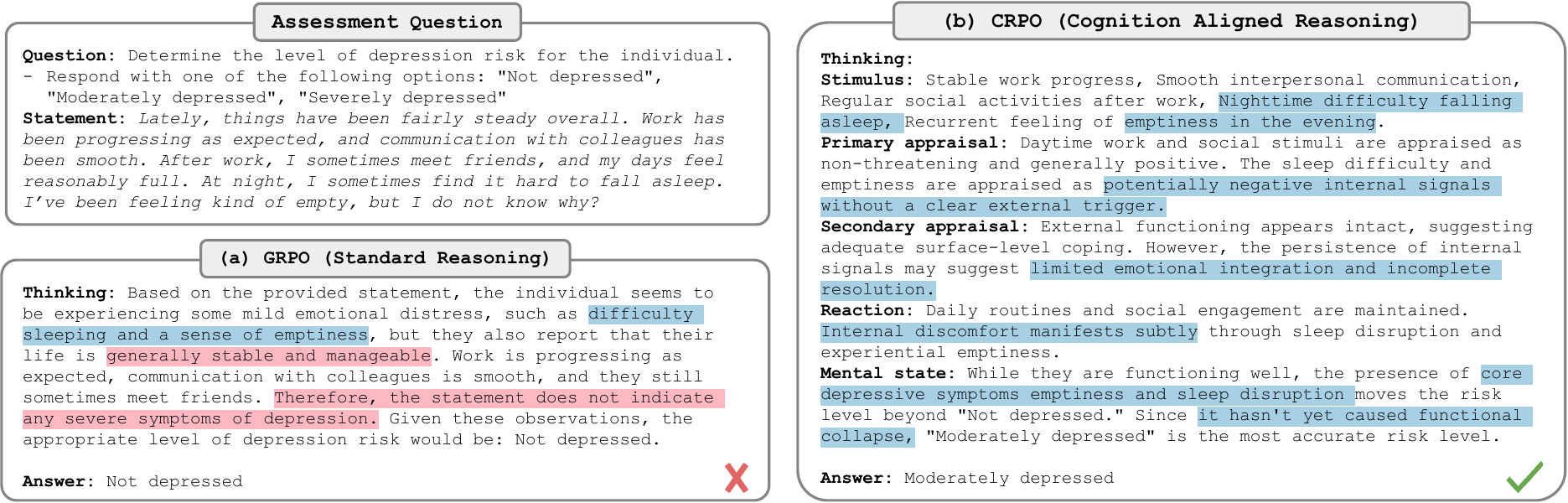}
  \caption{Comparison of reasoning outputs. (a) Standard reasoning produced by a GRPO-trained model. (b) Cognition-aligned reasoning generated by CRPO-trained Mental-R1, which correctly predicts the annotated depression risk.}
  \label{fig:casestudy}
\end{figure*}
\subsection{Q3: Reasoning Ability}
\label{section:reasoning}
We use GPT-5 as an external evaluator to score the reasoning demand of each test sample and retain the top 20\% within each of the eight datasets. This evaluation focuses on samples judged to require more complex reasoning. As shown in Figure~\ref{fig:reasoning_bars}, Mental-R1 consistently outperforms all baselines by a substantial margin, achieving an average improvement of approximately 15.6 weighted F1 points over the strongest baseline, with particularly large gains on DepSeverity and LID. These results support the effectiveness of CRPO on reasoning-intensive mental health assessment samples.

Figure~\ref{fig:casestudy} illustrates how CRPO reshapes reasoning compared with standard GRPO. The input describes stable external functioning alongside subtle internal cues, including sleep disturbance and emptiness. GRPO places more emphasis on positive external functioning and predicts the risk ``Not depressed.'' In contrast, CRPO initially considers both external functioning and internal signals, then progressively focuses on the  internal cues. By distinguishing preserved external functioning from unresolved internal distress, it arrives at the risk prediction ``Moderately depressed.'' This example illustrates how the cognitive stages organize competing cues and is consistent with the intended reasoning progression: considering multiple interpretations early and consolidating relevant evidence before reaching a final assessment.

\begin{figure*}[]
  \centering
\includegraphics[width=\textwidth]{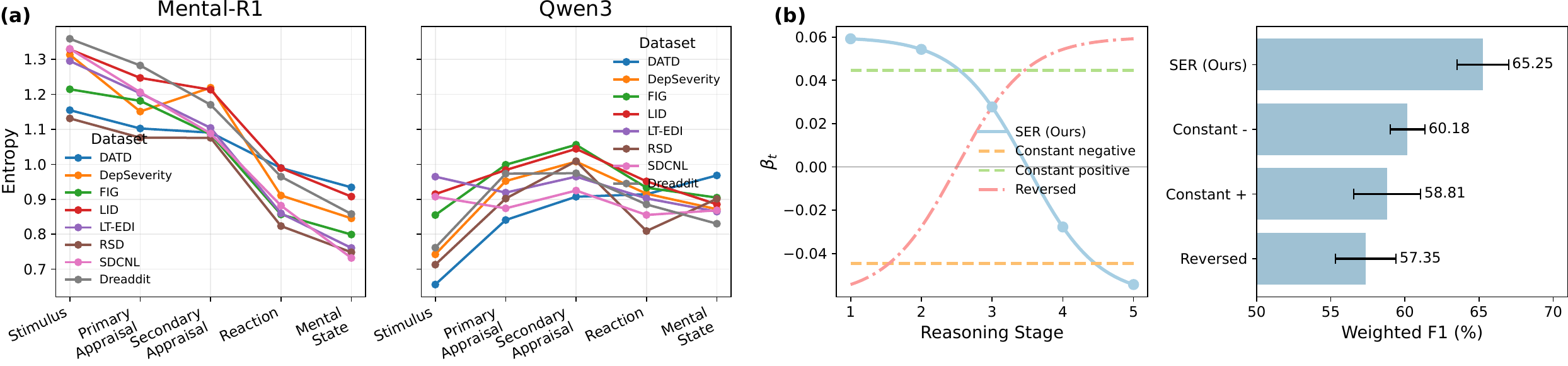}
  \caption{Analysis of SER. (a) Comparison of reasoning entropy across stages. We plot the average entropy for each reasoning stage across all test sets. Compared with vanilla Qwen3, CRPO-trained Mental-R1 shows a clear downward trend. (b) Visualization of uncertainty scheduling strategies and their corresponding performance. Left: different scheduling patterns across reasoning stages. Right: average weighted F1-score with standard deviation. The proposed SER follows a smooth transition from exploration to confidence and achieves the best performance. 
  }
  \label{fig:two}
\end{figure*}
\subsection{Q4: Entropy Dynamics and Scheduling in SER}
To examine whether SER produces the intended uncertainty dynamics, we compare the average token entropy at each reasoning stage for CRPO-trained Mental-R1 and the vanilla Qwen3 base model across all eight test sets. As shown in Figure~\ref{fig:two}(a), the entropy profile of Mental-R1 exhibits a gradual decline as the reasoning progresses, whereas the vanilla model shows relatively flat distribution throughout. This pattern is consistent with SER's intended progression from greater uncertainty in early stages toward more confident predictions in later stages.

We further compare our schedule with constant positive, constant negative, and reversed schedules (Figure~\ref{fig:two}(b)). Constant positive and negative coefficients encourage higher and lower entropy throughout reasoning, respectively. For a controlled comparison, their magnitudes equal the mean absolute value of $\beta_t$, while the reversed schedule uses the same coefficients in reverse order. Our schedule achieves the highest average weighted F1. Its advantage over constant schedules supports varying entropy regularization across stages. The performance decline under the reversed schedule further suggests that the direction of this variation matters, supporting the design of encouraging exploration early and reducing uncertainty as reasoning converges.
\section{Conclusion}
In this paper, we propose Cognitive Relative Policy Optimization (CRPO) to align large language model reasoning with real-world mental health assessment practice. CRPO combines stage-wise entropy regularization with theory-grounded reasoning stages to encourage early-stage exploration and late-stage certainty, mirroring the human cognitive progression from uncertainty to certainty. 
Across eight benchmark datasets, CRPO consistently outperforms multiple reinforcement learning baselines, and its trained model, Mental-R1, further surpasses strong LLM baselines, especially on reasoning-intensive samples. 
More broadly, this work suggests that incorporating cognition-aware uncertainty optimization can substantially improve model performance, offering a promising direction for developing human-aligned large reasoning models.



\bibliography{main}
\bibliographystyle{iclr2027_conference}
\clearpage
\appendix
\section{An Illustrative Example of Cognitive Appraisal Theory}
\label{app:cognitive_appraisal_example}
For example, consider public speaking as a stimulus. In the primary appraisal, the individual may perceive it as threatening due to potential embarrassment or negative evaluation. During the secondary appraisal, they evaluate their coping resources and abilities, such as preparation, prior experience, personal skills, or audience support. If these are judged insufficient, the reaction may involve anxiety, fear, or avoidance; if viewed as adequate, the response may instead involve confidence, focus, or even excitement. This appraisal process shows the process of human inner cognition when facing potential triggers in daily life.

\section{Dataset Details}
Stress state prediction involves determining whether an individual is experiencing stress: Dreaddit~\citep{turcan2019dreaddit}. Anxiety state prediction aims to identify whether a person expresses signs of anxiety or depression: DATD~\citep{owen2020towards}. Depression level classification assigns individuals to non-depressed, moderately depressed, or severely depressed categories: LT-EDI~\citep{sampath2022data}. Depression severity prediction further refines this by classifying individuals into minimal, mild, moderate, or severe cases:  DepSeverity~\citep{naseem2022early}. Note that DepSeverity is built upon the same textual corpus as Dreaddit but annotated for a different task. Suicidal ideation detection predicts whether an individual is at risk of suicide: SDCNL~\citep{haque2021deep}. Suicide risk severity classification assesses individuals across four levels of suicide risk, including indicator, ideation, behavior, and attempt: RSD~\citep{zheng2025rsd}. For RSD, we retain only the first post encountered for each individual in the original dataset, as multiple posts from the same individual may share the same risk label and contain overlapping content. 
Loneliness state prediction identifies whether an individual is experiencing loneliness: FIG~\citep{jiang2022many}.
Loneliness intensity prediction categorizes an individual's loneliness into four ordered intervals, representing increasing levels of loneliness severity: LID~\citep{LID}.

\begin{algorithm}[t]
\caption{Cognitive Relative Policy Optimization (CRPO)}
\label{alg:crpo}
\small
\KwIn{
Training set $\mathcal{D}$; 
current policy $\pi_{\theta}$; 
sampling policy $\pi_{\theta_{\text{old}}}$; 
reference policy $\pi_{\text{ref}}$; 
group size $G$; 
cognitive stages $\{c_t\}_{t=1}^{S}$; 
entropy schedule parameters $M$ and $\tau$.
}
\KwOut{Optimized policy $\pi_{\theta}$.}

\For{each training iteration}{
    Sample a mini-batch of prompts $\{q\}$ from $\mathcal{D}$\;
    
    \For{each prompt $q$}{
        Sample a group of outputs $\{o_1, \dots, o_G\} \sim \pi_{\theta_{\text{old}}}(\cdot \mid q)$\;
        
        \For{each output $o_i$}{
            Compute reward $r_i = r_f(o_i) + \eta_{d,y}r_a(o_i, y)$ using format and answer rewards\;
            
            Segment $o_i$ into cognitive stages $\{c_t\}_{t=1}^{S}$ via stage tags\;
            
            \For{each stage $t$}{
                Identify token indices $N_{i,t}$ corresponding to stage $c_t$\;
                
                Compute token entropy $\mathcal{H}_{i,k}(\theta)$ for each $k \in N_{i,t}$\;
                
                Compute stage-level entropy
                $
                \bar{\mathcal{H}}_{i,t}(\theta) = \frac{1}{|N_{i,t}|} \sum_{k \in N_{i,t}} \mathcal{H}_{i,k}(\theta)
                $\;
            }
        }
        
        Compute advantages
        $
        A_i = \frac{r_i - \text{mean}(\{r_j\}_{j=1}^{G})}{\text{std}(\{r_j\}_{j=1}^{G})}
        $\;
    }
    
    Compute stage-wise entropy coefficients
    $
    \beta_t=-M \cdot \frac{e^{t-\tau}-e^{-(t-\tau)}}{e^{t-\tau}+e^{-(t-\tau)}}, \quad t = 1, \dots, S
    $\;
    
    Compute entropy regularization
    $
    \mathcal{J}_{\mathrm{SER}}(\theta) = \frac{1}{G} \sum_{i=1}^{G} \sum_{t=1}^{S} \beta_t \bar{\mathcal{H}}_{i,t}(\theta)
    $\;
    
    Compute the objective $\mathcal{J}_{\mathrm{GRPO}}(\theta)$ using $\rho_{i,k}(\theta)$ and $A_i$\;
    
    Update $\theta$ by maximizing
    $
    \mathcal{J}_{\mathrm{CRPO}}(\theta) = \mathcal{J}_{\mathrm{GRPO}}(\theta) + \mathcal{J}_{\mathrm{SER}}(\theta)
    $\;
}
\Return{$\pi_{\theta}$}
\end{algorithm}

\section{Experiment Details}
\label{app:details}
The learning rate is set to $5\times10^{-6}$. The SER parameters $M$ and $\tau$ are fixed to 0.06 and 3.5, respectively, the KL coefficient is set to 0.01, and the format reward magnitude $\alpha$ is set to 0.5. Training is performed for one epoch with a per-device batch size of 4 and gradient accumulation over 16 steps. We use 4 generations per prompt, with both the maximum prompt length and maximum completion length set to 512 tokens. All experiments are conducted in bf16 precision with DeepSpeed ZeRO-2 optimization. For all RL-based methods, we use Qwen3-8B~\citep{yang2025qwen3} as the base model. The model is jointly trained and validated on the combined training and validation splits of all datasets, and then evaluated separately on each dataset’s test split. All experiments are conducted on a Linux server equipped with eight NVIDIA RTX PRO 6000 GPUs. Our framework is implemented using PyTorch, Transformers, and TRL.
Following~\citep{yang2024mentallama,zhai2025mentalglm}, we evaluate model performance with weighted F1-score. We report the mean and standard deviation over five independent inference.

We follow the standard train/validation/test splits provided in the original datasets. For datasets without predefined splits, we randomly divide the data into training, validation, and test sets with a ratio of 8:1:1. For datasets that only provide training and test sets, we further randomly split 10\% of the training data as a validation set. As an exception, we align the partitions of DepSeverity with Dreaddit's splits due to their shared corpus.

\subsection{Baseline and Ablation Configurations}

The RL baselines, including GRPO and DAPO, use free-form
reasoning with a standard \texttt{<think>}/\texttt{<answer>}
format reward and an answer-correctness reward, without
SER and cognitive-stage constraints. 

The \textit{w/o SER} variant retains the cognitive-stage
instructions and reward design of CRPO, while removing the
entropy regularization term. 
The \textit{w/o CRS} variant
removes the cognitive-stage instructions and the corresponding
stage-specific format checks, while retaining the outer output
structure and the answer reward used in full CRPO. Its reasoning
tokens within \texttt{<think>} are divided into $S=5$ contiguous
segments of approximately equal length, and SER is applied to
these segments using the same coefficient schedule as full CRPO,
with $M=0.06$ and $\tau=3.5$.

\begin{table*}[t]
\centering
\scriptsize
\caption{Summary of the eight human-annotated datasets used in this study. The datasets span multiple mental disorders and task formulations, providing a comprehensive benchmark for evaluating performance on mental health assessment.}
\label{tab:datasets}
\setlength{\tabcolsep}{5pt}
\begin{tabular}{l l l c c p{3.9cm}}
\toprule
\textbf{Dataset} & \textbf{Disorder} & \textbf{Task} & \textbf{\#Samples} & \textbf{\#Classes} & \textbf{Label Space} \\
\midrule

Dreaddit 
& Stress 
& State prediction 
& 3,553 & 2 
& No, Yes \\

DATD 
& Anxiety 
& State prediction 
& 1,050 & 2 
& No, Yes \\

LT-EDI 
& Depression 
& Level classification 
& 10,251 & 3 
& Not, Moderate, Severe \\

DepSeverity 
& Depression 
& Severity prediction 
& 3,553 & 4 
& Minimum, Mild, Moderate, Severe \\

SDCNL
& Suicide 
& Ideation detection 
& 1,895 & 2 
& No, Yes \\

RSD 
& Suicide 
& Severity classification 
& 1,265 & 4 
& Indicator, Ideation, Behavior, Attempt \\

LID 
& Loneliness 
& Intensity prediction 
& 498 & 4 
& [1--2], [2--3], [3--4], [4--5] \\

FIG 
& Loneliness 
& State prediction 
& 5,633 & 2 
& No, Yes \\

\bottomrule
\end{tabular}
\end{table*}

\begin{wrapfigure}{r}{0.3\columnwidth}
  \centering
  \vspace{-20pt}
\includegraphics[width=\linewidth]{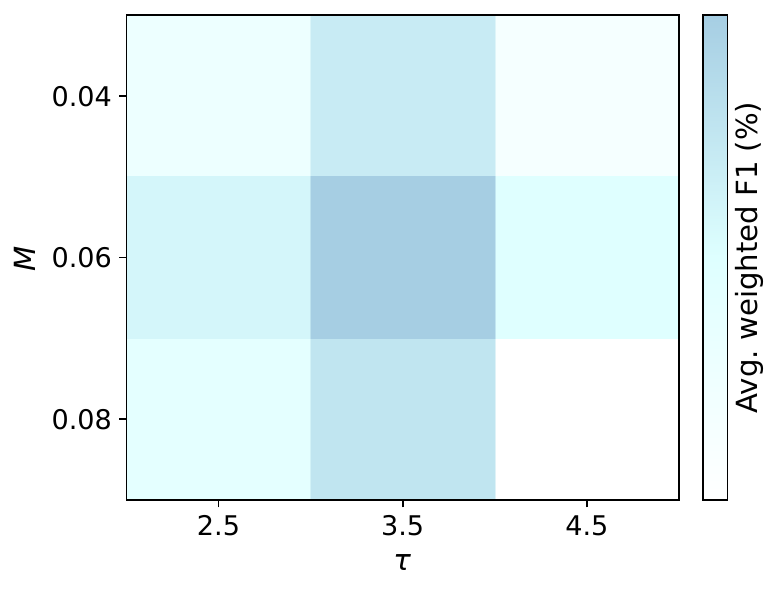}
  \caption{Heatmap of average weighted F1-score under different combinations of the magnitude $M$ and transition parameter $\tau$. Darker color indicates better performance.
  }
  \label{fig:parameter}
  \vspace{-10pt}
\end{wrapfigure}
\section{Parameter Study of SER}
We analyze the influence of the magnitude parameter $M$ and the transition parameter $\tau$ in SER. Specifically, $M$ controls the overall strength of entropy modulation, while $\tau$ determines the stage at which the schedule shifts from exploration to certainty. We evaluate a grid with $M \in \{0.04, 0.06, 0.08\}$ and $\tau \in \{2.5, 3.5, 4.5\}$, and report the results in Figure~\ref{fig:parameter}. 
As shown, performance peaks at a moderate magnitude $M=0.06$ and decreases only slightly when the modulation is either weaker or stronger, indicating that SER is relatively insensitive to $M$ within a moderate range. In contrast, performance is consistently highest when $\tau=3.5$, compared with earlier transitions at $\tau=2.5$ and later transitions at $\tau=4.5$. This suggests that shifting to certainty too early limits necessary exploration, while delaying the transition weakens decisiveness in later stages. The best-performing region therefore corresponds to a mid-stage transition after the second appraisal stage, allowing the model to maintain exploration throughout the context-gathering and appraisal phases before pivoting to certainty for the final reaction and mental state prediction.

\section{System Prompt for Mental-R1}
\label{app:cog_prompt}

The prompt in last line will be replaced with the specific mental health question during usage.

\textit{``A conversation between User and Assistant. The user asks a question, and the Assistant solves it. The Assistant must explicitly think through the reasoning process before giving the final answer.}

\textit{The reasoning process should strictly follow the stages below:}

\textit{1. \textless{}stimulus\textgreater{} Identify the key event, situation, or object described by the user without interpretation. \textless{}/stimulus\textgreater{}}

\textit{2. \textless{}primary\_appraisal\textgreater{} Assess the personal relevance and potential impact of the stimulus. \textless{}/primary\_appraisal\textgreater{}}

\textit{3. \textless{}secondary\_appraisal\textgreater{} Evaluate available resources and options for coping with the situation. \textless{}/secondary\_appraisal\textgreater{}}

\textit{4. \textless{}reaction\textgreater{} Describe the likely affective and behavioral responses. \textless{}/reaction\textgreater{}}

\textit{5. \textless{}mental\_state\textgreater{} Conclude the probable mental health condition or state. \textless{}/mental\_state\textgreater{}}

\textit{The reasoning process and answer must be enclosed within \textless{}think\textgreater{} \textless{}/think\textgreater{} and \textless{}answer\textgreater{} \textless{}/answer\textgreater{} tags respectively, i.e.:}

\textit{\textless{}think\textgreater{}}

\textit{\textless{}stimulus\textgreater{} stimulus here \textless{}/stimulus\textgreater{}}

\textit{\textless{}primary\_appraisal\textgreater{} primary appraisal here \textless{}/primary\_appraisal\textgreater{}}

\textit{\textless{}secondary\_appraisal\textgreater{} secondary appraisal here \textless{}/secondary\_appraisal\textgreater{}}

\textit{\textless{}reaction\textgreater{} reaction here \textless{}/reaction\textgreater{}}

\textit{\textless{}mental\_state\textgreater{} mental state here \textless{}/mental\_state\textgreater{}}

\textit{\textless{}/think\textgreater{}}

\textit{\textless{}answer\textgreater{} answer here \textless{}/answer\textgreater{}}

\textit{User: prompt. Assistant:''}

\section{Prompt Templates for Different Datasets}

To ensure consistent evaluation across datasets with heterogeneous annotation schemes, we employ dataset-specific prompt templates that align with each task’s original label space. All prompts follow a unified instruction--response format.

{Binary Classification Tasks.}
For datasets formulated as binary classification, the model is instructed to respond with \textit{Yes} or \textit{No}.

\begin{itemize}[leftmargin=*]
  \item DATD (Anxiety/Depression Risk):  
  \textit{``Determine whether the individual who made the following statement is at risk of anxiety or depression.  
  Respond with either `Yes' or `No'.  
  Statement:''}

  \item Dreaddit (Psychological Stress):  
  \textit{``Determine whether the individual who made the following statement is experiencing psychological stress.  
  Respond with either `Yes' or `No'.  
  Statement:''}

  \item FIG (Loneliness Detection):  
  \textit{``Determine whether the individual who made the following statement is experiencing loneliness.  
  Respond with either `Yes' or `No'.  
  Statement:''}

  \item SDCNL (Suicide Risk Detection):  
  \textit{``Determine whether the individual who made the following statement is at risk of suicide.  
  Respond with either `Yes' or `No'.  
  Statement:''}
\end{itemize}

{Multi-Class Severity Classification Tasks.}
For datasets requiring ordinal or categorical severity prediction, the prompt explicitly enumerates all valid label options.

\begin{itemize}[leftmargin=*]
  \item DepSeverity (Depression Severity):  
  \textit{``Determine the level of depression risk for the individual who made the following statement.  
  Respond with one of the following options: `Minimum', `Mild', `Moderate', `Severe'.  
  Statement:''}

  \item LT-EDI (Depression Risk Level):  
  \textit{``Determine the level of depression risk for the individual who made the following statement.  
  Respond with one of the following options: `Not depressed', `Moderately depressed', `Severely depressed'.  
  Statement:''}

  \item RSD (Suicide Risk Severity):  
  \textit{``Determine the level of suicide-related content for the individual who made the following statement.  
  Respond with one of the following options: `Indicator', `Ideation', `Behavior', `Attempt'.  
  Statement:''}

  \item LID (Loneliness Intensity):  
  \textit{``Determine the intensity of loneliness for the individual who made the following statement.  
  Respond with one of the following options: `[1--2]', `[2--3]', `[3--4]', `[4--5]'.  
  Note that 1 indicates not lonely and 5 indicates the highest intensity.  
  Statement:''}
\end{itemize}

\section{Prompt for Reasoning-Focused Subset Construction}
\label{app:reasoning_subset}
\textit{
You are curating items that genuinely REQUIRE reasoning to infer
a mental-health label.\\[0.5em]
Dataset purpose: \{dataset\_purpose\}\\
Label set: \{label\_set\}\\
Text:\\
\textless{}\textless{}\textless{}\{text\}\textgreater{}\textgreater{}\textgreater{}\\[0.5em]
Score the degree to which CORRECTLY inferring the gold label
requires nontrivial reasoning beyond explicit keywords or
self-disclosure.\\[0.5em]
Guidelines:\\
- Penalize explicit self-report (e.g., ``I have depression'',
``I'm suicidal'') and single-word shortcuts.\\
- Reward implicit, multi-clue, temporal/causal, contrastive cues
and so on (no need to enumerate axes).\\[0.5em]
Return ONLY JSON:\\
\{\\
\hspace*{1em}"notes": "reason",\\
\hspace*{1em}"reasoning\_need\_score": 0-10,\\
\}
}
\end{document}